\documentclass{article} 
\usepackage[final]{colm2026_conference}

\usepackage{microtype}
\usepackage{hyperref}
\usepackage{url}
\usepackage{booktabs}
\usepackage{hyperref}

\usepackage{graphicx}
\usepackage{subcaption}  

\usepackage{amsmath}
\usepackage{amssymb}
\usepackage{mathtools}
\usepackage{amsthm}

\usepackage{listings}
\usepackage{booktabs}
\usepackage{tabularx}
\usepackage{ragged2e}
\usepackage{makecell}
\usepackage{wrapfig}

\usepackage{xcolor}

\usepackage{soul}

\usepackage[capitalize,noabbrev]{cleveref}

\usepackage[T1]{fontenc}

\theoremstyle{plain}

\theoremstyle{definition}

\theoremstyle{remark}

\usepackage[textsize=tiny]{todonotes}

\usepackage{lineno}

\definecolor{darkblue}{rgb}{0, 0, 0.5}
\hypersetup{colorlinks=true, citecolor=darkblue, linkcolor=darkblue, urlcolor=darkblue}

\definecolor{lightgreen}{HTML}{b8e0c4}
\newcommand{\hlgreen}[1]{{\sethlcolor{lightgreen}\hl{#1}}}

\title{Will Scaling Improve Social Simulation with LLMs?}

\author{Caleb Ziems\textsuperscript{1}, William Held\textsuperscript{1,2}, Su Doga Karaca\textsuperscript{1}, \\ 
\textbf{David Grusky\textsuperscript{1}, Tatsunori Hashimoto\textsuperscript{1}, Diyi Yang\textsuperscript{1}} \\
\textsuperscript{1}Stanford University, \textsuperscript{2}Open Athena \\
\texttt{\{cziems,held,sukaraca,grusky,thashim,diyiy\}@stanford.edu}
}

\begin{document}

\ifcolmsubmission
\linenumbers
\fi

\maketitle
\begin{abstract}
Large Language Model (LLM) social simulations are a promising research method, but they are not yet faithful enough to be adopted widely. In this work, we investigate whether the current scaling paradigm in language modeling is likely to close these gaps, or whether simulation fidelity is orthogonal to general capabilities and therefore deserving of more research attention. We use scaling laws to study the relationship between LLMs' compute scale, general capability benchmarks, and the fidelity of social simulation in three representative sub-domains: opinion modeling, behavioral simulation, and longitudinal forecasting. Surprisingly, we discover strong compute scaling in all three settings, using a suite of 85 transformer LLMs with the Qwen3 architecture pre-trained on the DCLM web text corpus under fixed-compute budgets from $10^{18}$ to $10^{20}$ FLOPs. Then we evaluate 35 larger and more capable open-weight models up to 70B parameters, allowing us to predict downstream accuracy from loss. This reveals that the majority of behavioral and opinion simulation tasks will rapidly improve with scale, particularly when they involve populations that are well-represented in English web corpora. Longitudinal forecasting and underrepresented opinions scale more slowly, especially when they are less correlated with general knowledge and reasoning benchmarks like MMLU. In behavior simulation, scaling fails to improve model calibration with human cognitive biases like risk aversion, as well as human heuristics like learning correlated rewards from related tasks. On these tasks, even fine-tuned models fail to noticeably scale up performance from 0.5B to 8B parameters. Taken together, we conclude that scale will improve social simulations in most settings, but outliers exist, and improvements will be less reliable in low-resource domains.
\end{abstract}

\section{Introduction}
\label{sec:intro}
Scaling laws are a useful method for anticipating LLMs’ future capabilities as models scale up \citep{kaplan2020scaling}. For tasks like question-answering, which are explicitly targeted during model development and have clear ground truth, performance has monotonically increased with model size \citep{myrzakhan2024open}. Conversely, social simulation is not privileged in LLM training decisions, and instead requires models to match distributions of opinions, stochastic behaviors, and sequential dependencies \citep{park2024generative,kolluri2025finetuning,salganik2020measuring}. Generally, LLMs are not well-calibrated for such tasks \citep{bisbee2024synthetic,durmus2023towards,hu2025simbench,sorensen2025spectrum,tan2024language}. The existence and directionality of scaling laws here are unknown. If we can show the current trajectory in language modeling will rapidly improve LLMs' simulation fidelity, then researchers will have strong cost and time incentives to begin utilizing LLMs for pilot studies, sensitivity analysis, and replications at scale \citep{anthis2025llm}. But if simulations don't scale, or require distinct capabilities orthogonal to reasoning and math, then our findings will motivate more concerted research on faithful LLM simulations.

In this work, we specifically look at how LLMs’ pre-training compute and general benchmarking abilities relate to their ability to simulate social variables over \textit{finite} probability spaces (e.g., multiple-choice) where likelihood-based scaling is well-defined \citep{kaplan2020scaling,hoffmann2022training}. Future work can investigate scaling properties of open-ended generation and agentic systems \citep{park2022social,wang2025user,taubenfeld2024systematic,mou2024individual,tranchero2024theorizing,weidmann2025measuring} where the target object is a high-level semantic property of entire sequences or agentic interactions. Finite probability spaces are the natural setting for a large subset of the simulation literature, including work on opinion surveys \citep{suh2025language}, lab experiments \citep{chen2025manager}, economic games \citep{del2025can}, consumer surveys \citep{brynjolfsson2025augmenting}, and longitudinal studies \citep{salganik2020measuring}. Across these numerous application domains, we focus on three of the most common inference problems:

\begin{enumerate}
    \item \textbf{Survey Simulation} $P(y|x)$: modeling survey responses $y$ for a given survey context $x$, including demographic information and the survey question.
    \item \textbf{Behavioral Simulation} $P(a_t|h_t)$: modeling actions $a_t$ conditioned on history $h_t = (a_1, r_1, ..., a_{t-1}, r_{t-1})$ composed of prior actions $a_i$ and reward signals $r_i$.
    \item \textbf{Longitudinal Forecasting} $P(y_{t+k}|y_t,x)$: predicting future social variables $y_{t+k}$ from past social variables $y_t$ and social context $x$.
\end{enumerate}

We operationalize the three inference problems with three tasks: the \textit{World Values Survey} (\citeyear{wvs2022wave7}), \textit{Psych-101} \citep{binz2025foundation}, and the \textit{Americans' Changing Lives} \citep{acl2024dataset} study. First, we establish compute scaling laws on these tasks by evaluating 85 transformer LLMs from 0.2B to 12B parameters, varying the compute-budget from $10^{18}$ to $10^{20}$ FLOPs, but fixing all other model details like architecture and pre-training distribution. We find surprisingly clean log-linear scaling laws explain as much 97\% of the variance in loss (Figure~\ref{fig:compute_scaling}). However, computational constraints generally limit such controlled scaling suites to smaller models below 12B parameters \citep{mcleish2026gemstones,held2025relative,bhagia2025establishing,biderman2023pythia}. To understand the implications for downstream utility with larger, more capable models, we establish calibration functions from observational experiments with 35 open-weight models up to 70B parameters, allowing us to predict accuracy from loss. 

We find that scaling up LLMs by 1-2 orders of magnitude significantly improves performance on a broad range of opinion and behavior simulation tasks. However, for longitudinal forecasting, and a subset of decision-making tasks, accuracy does not scale well. Moreover, only some social simulation results are correlated with general model competence (the Pearson $r$ ranges from 0 to 0.85), with particularly weak correlations for demographic populations that are underrepresented in pre-training data. Moreover, scaling fails to improve model calibration with human cognitive biases like risk aversion, as well as human heuristics like learning correlated rewards from related tasks. To challenge our null scaling conclusions, we fine-tune a suite of Llama3 and Qwen2.5 models from 0.5B to 8B parameters. However, on these null scaling tasks, we also fail to observe parameter scaling after fine-tuning. This shows that, although aggregate simulation will improve with general model improvement, outliers exist, and improvements will be less reliable in low-resource domains.

To sum up, our contributions have four components.(\S\ref{sec:tasks})~\textbf{Social simulation tasks:} we operationalize three benchmarks, along with one new dataset, for scaling law analysis. (\S\ref{sec:compute_scaling_laws})~\textbf{Compute scaling laws:} we carefully measure the impact of scaling compute on task loss for social simulation.
(\S\ref{sec:obs_scaling_laws})~\textbf{Observational scaling laws:} we conduct large-scale observational scaling law experiments across diverse model families, quantifying the relationship between general benchmark performance and social simulation fidelity. And
(\S\ref{sec:finetuning})~\textbf{Trained models:} we release fine-tuned variants of Llama3 and Qwen2.5 models trained on our tasks, along with all datasets and evaluation code, to support future research on socially faithful language modeling here: \url{https://github.com/SALT-NLP/social-scaling}.

\newpage

\section{Related Work}
\label{sec:related_work}

\textbf{LLM Utility in the Social Sciences.} Qualitative studies show that social scientists are eager yet apprehensive to use LLMs as tools \citep{fecher2025friend,schroeder2025qualitative,morris2023scientists}. Already, LLMs can reliably assist with literature review \citep{liao2024research}, data labeling \citep{ziems2024css,dai2023llm}, survey design \citep{bail2024generative}, programming \citep{jimenez2024swebench}, and paper writing \citep{liao2024research} without domain-specific training. As predictive models, they can forecast average treatment effects for a wide range of social science studies \citep{hewitt2024predicting}. Fine-tuning may further enable LLMs to generate new hypotheses \citep{bazgir2025agentic} and research ideas \citep{si2026towards}. SFT helps models better predict public opinion \citep{suh2025language} and human behavior, both in the lab \citep{binz2025foundation} and in the wild \citep{kolluri2025finetuning}. And with enough in-context data, LLMs can simulate individuals' opinions, personalities, and decision-making behaviors at high levels of fidelity \citep{park2024generative}. Together, these results suggest that LLMs will be increasingly integrated into the social science pipeline at all stages.

\textbf{Social Simulation.} Many prior studies evaluate LLMs as simulators of social variables over \textit{finite} probability spaces (e.g., multiple-choice). For example, LLMs have been used to simulate human opinion and value surveys \citep{abdurahman2024perils,argyle2023out,bisbee2024synthetic,boelaert2025machine,dominguez2024questioning,jiang2024donald,kim2023ai,qu2024performance,santurkar2023whose,suh2025language}, psychological experiments \citep{binz2025foundation,chen2025manager,liu2024large,park2024diminished}, economic games \citep{aher2023using,horton2023large,del2025can,gonzalez2025llms,jia2024can,ross2024llm}, consumer surveys \citep{brand2023using,brynjolfsson2025augmenting}, HCI user studies \citep{hamalainen2023evaluating}, and aggregate experimental outcomes \citep{hewitt2024predicting,manning2024automated}, both synchronically and longitudinally \citep{ahnert2025extracting}. Studies have also evaluated LLM agents with psychometric inventories \citep{park2024generative,pellert2024ai,petrov2024limited,wang2024not} and typicality ratings \citep{heyman2024impact}. In these settings, tasks can be formatted as multiple-choice, which is suitable for likelihood-based scaling analysis. Other work on the simulation of open-ended generation and agentic systems is less readily suited for scaling analysis. In these settings, researchers typically rely on human judges to rate the LLM's human-likeness in simulated social environments \citep{park2022social,wang2025user}, including dyadic debates \citep{taubenfeld2024systematic}, social media platforms \citep{yang20oasis}, customer service \citep{zhou2026mind}, multi-agent collaborations \citep{tranchero2024theorizing,weidmann2025measuring}, and small-scale societies \citep{mou2024individual}. Across both lines of work, the scaling behavior of simulation fidelity remains to be studied in depth. 

\textbf{Social Scaling Laws.} We are the first to study social simulation performance with parameter-controlled compute-scaling laws. Related to this work, \citet{held2025relative} evaluate careful parameter-controlled IsoFLOP sweeps to measure disparities in the scaling of dialect robustness and AI risk behaviors. \citet{hu2025simbench} provide an observational analysis of behavioral simulation across 45 models and find a roughly log-linear scaling trend in smaller models with diminishing returns in larger models. \citet{merali2024scaling} fit an economic scaling law that shows how, in human-AI partnerships, workers' earnings are a log-linear function of LLM training compute. And \citet{hackenburg2025scaling} observe starkly diminishing returns in LLM persuasiveness with scale.

\section{Social Simulation Tasks}
\label{sec:tasks}
The motivation for this work is to anticipate the future utility of LLMs for simulating experiments in the social sciences. In particular, we consider \textbf{opinion surveys}, \textbf{behavioral experiments}, and \textbf{longitudinal studies} as three representative domains in which LLMs simulations could transform social science but haven't because even today's best models lack sufficient fidelity. Concretely, opinion simulation requires LLMs to accurately model opinion \textit{distributions} rather than a single ground-truth answer. Even the best LLMs collapse variance and heterogeneity in opinion or value distributions  \citep{bisbee2024synthetic,durmus2023towards}, and this is exacerbated in post-training \citep{hu2025simbench,sorensen2025spectrum}. Relatedly, behavioral simulation requires approximating stochastic human choices, but LLMs often violate cognitive constraints on behavior \citep{bowers2025centaur,gao2024take}. Finally, longitudinal simulation and forecasting requires modeling temporal dynamics, but LLMs are weak models of time series and sequential dependencies \citep{tan2024language}. As such, longitudinal simulation tasks remain unsolved \citep{salganik2020measuring}.

The above represent the leading challenges in single-agent simulations for the social sciences. There are other high-impact settings involving multi-agent simulations, but these introduce additional design layers in the agent-interaction protocol which can complicate scaling analyses. As the first study of its kind, we focus on single-agent, MCQ-style simulation tasks. We will now briefly discuss how we operationalize and construct each task. Details regarding the precise prompts, inputs, and outputs for this task are in Appendix~\ref{appdx:prompts}.

\paragraph{Simulating Opinions: \texttt{WVS}.} We use data from the largest and most comprehensive set of global opinion surveys to date, the 7th wave of the \citet[WVS;][]{wvs2022wave7}. Following \citet{durmus2023towards}, we evaluate LLMs' ability to model the opinion distribution for a given target demographic group. Unlike \citet{durmus2023towards}, which targets only high-level national cultures, we define our target groups at the intersection of 7 explanatory demographic variables: (1) the country or region of origin, (2) sex, (3) age, (4) education, (5) socioeconomic class, (6) religiosity, and (7) whether the participant is from a rural or urban area. For each target demographic, we aggregate the WVS responses to create an opinion distribution and ensure that questions have at least 30 responses from the target demographic. This leaves data from the following WVS survey units: Bangladesh, Bolivia, Canada, China, Egypt, Hong Kong, India, Indonesia, Libya, Myanmar, Nigeria, Pakistan, South Korea, Thailand, Tunisia, Turkey, and Vietnam. Prompts take the form of multiple-choice questions (see Appendix~\ref{appdx:prompt_wvs}).

\paragraph{Simulating Behavior: \texttt{Psych-101}} We use Psych-101 \citep{binz2025foundation}, since this is one of the largest and most diverse collections of psychological experiments available. Psych-101 transcribes experiments with discrete action spaces, indicated by alphabetical indices, and we compute model loss on these indices (see Appendix~\ref{appdx:prompt_psych101}). Most of these experiments are reward maximization tasks like single- and multi-task bandits (44\%). Other task designs include associative learning experiments (10\%), and experiments which cover cognitive heuristics and decision-making (23\%).

\paragraph{Forecasting Longitudinal Studies: \texttt{ACL}} We are aware of only one shared task for predicting longitudinal study results known as the Fragile Families Challenge \citep{salganik2020measuring}, focused on the life outcomes of at-risk children. However, we consider a more general domain and construct our own dataset from Americans’ Changing Lives \citep[ACL;][]{acl2024dataset}. The ACL is the longest-running nationally representative longitudinal panel survey designed to track how social, economic, and psychological factors shape individuals’ lives over time in the United States. The survey was collected in 6 waves, in the years 1986, 1989, 1994, 2001, 2011, and 2019. We set up the ACL task as next wave prediction, where the output is the predicted answer of a respondent to a target question in the 2019 wave and the input includes the values of all explanatory, control, and target outcome variables from prior waves (see Appendix~\ref{appdx:prompt_acl}).

\section{Compute Scaling Laws for Social Simulation}
\label{sec:compute_scaling_laws}

Compute scaling laws follow the discovery that the loss $L_m$ of LLM $m$ on some held out set is inversely related to the model's scale, dataset size, and compute measures $C_m$, following predictable power law trends \citep{kaplan2020scaling} given by
\begin{equation*}
    \log(L_m) \approx \beta_f\log(C_m) + \alpha_f
\end{equation*}

The intercept $\alpha_f$ represents a hypothetically initial error, and the scaling coefficient $\beta_f$ indicates the rate of improvement on the task from scaling compute alone.

Here, we establish LLMs' compute-scaling behavior on our three social simulation tasks from \S\ref{sec:tasks}, using a suite\footnote{\url{https://huggingface.co/collections/marin-community/dclm-baseline-isoflop-models}} of carefully-controlled model checkpoints from \citet{held2025relative}. Specifically, we evaluate a suite of 85 decoder-only Transformers with the Qwen3 architecture~\citep{qwen3} trained on the DCLM pre-training corpus \citep{dclm} with fixed compute budgets $C_m$ ranging from $10^{18}$ to $10^{20}$ FLOPs. 

For each FLOP budget, we sweep token and model size to select the compute-optimal token count. Then, along the compute-optimal points, we report task loss as a function of compute (FLOPs). This IsoFLOP approach from \citet{hoffmann2022training} has been found to be the most stable method for fitting compute scaling laws.  

\begin{figure}
    \centering
    \includegraphics[width=\linewidth]{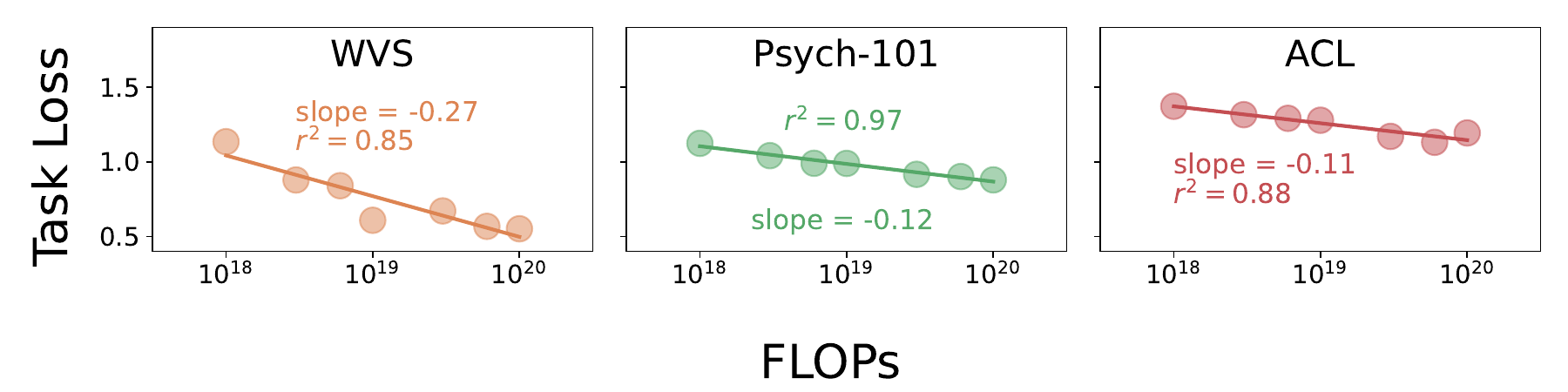}
    \caption{\small \textbf{Compute Scaling Laws.} We observe log-linear improvements in task loss on all three social science tasks after scaling compute alone with models trained on DCLM \citep{dclm} from $10^{18}$ to $10^{20}$ FLOPs.} 
    \label{fig:compute_scaling}
\end{figure}

\paragraph{Results.} Figure~\ref{fig:compute_scaling} shows log-linear scaling laws in all three social science tasks. Scaling compute alone consistently improves task performance, explaining at least 85\% of model variance in all three settings: behavior simulation (\texttt{Psych-101}, $r^2=0.97$), longitudinal forecasting (\texttt{ACL}, $r^2=0.88$), and opinion simulation (\texttt{WVS}, $r^2=0.85$). These scaling trends generally hold across the \texttt{WVS} and \texttt{Psych-101} and subtasks\footnote{ One outlier is in \texttt{Psych-101}, where the \citet{schulz2020finding} subtask has a slope of zero. We investigate this anomaly in \S\ref{sec:finetuning}.} as well (see Figure~\ref{fig:compute_scaling_psych101} in Appendix~\ref{appdx:compute_scaling_laws} for more details).

Overall, these results show that \textit{MCQ-style social simulation is a skill that can be learned directly from pre-training}. This is surprising. To some degree, these tasks require an understanding of social and psychological behaviors that are embedded in ambiguous cultural contexts, and these contexts are underrepresented in web corpora \citep{alkhamissi2024cultural}. Still, the skills to make sense of them derive from web pre-training corpora.

\begin{figure}
    \centering
    \includegraphics[width=0.9\linewidth]{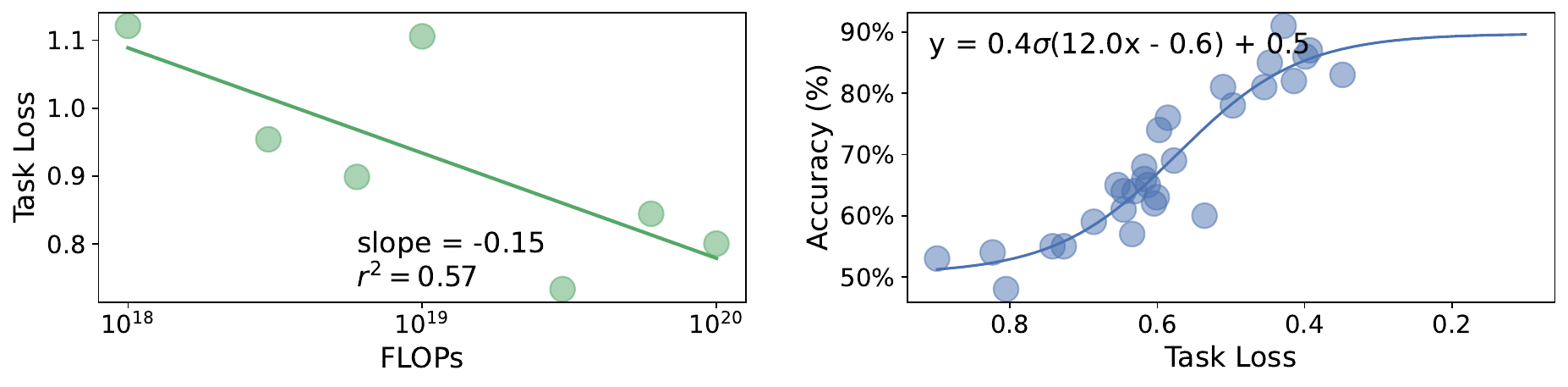}
    \caption{\small \textbf{Compute-optimal scaling and downstream forecasting.} \textbf{Left:} Compute scaling laws for \citet{hilbig2014generalized}, a subtask of \texttt{Psych-101}. \textbf{Right:} We show this loss correlates with accuracy sigmoidally, so loss can serve as a proxy for downstream progress.} 
    \label{fig:compute_scaling_psych101_subtask}
\end{figure}

Our findings so far seem to reinforce the \textit{bitter lesson} \citep{sutton2019bitter}, indicating that scale alone can produce useful models for social science simulation. However, before we can make conclusions about the utility of models, we need to establish a link between loss scaling and downstream metrics of interest. The checkpoints we use for compute scaling laws are not strong enough to reliably establish this link. For this reason, we instead adopt the \textit{two-stage} procedure of \citet{llama3}, shown in Figure~\ref{fig:compute_scaling_psych101_subtask}. After we have the compute-scaling law on task loss, we map loss to accuracy with a linear or sigmoidal calibration function. In \S\ref{sec:obs_scaling_laws}, we fit these calibration functions observationally \citep{ruan2024observational} to the performances of a wider range of open-weight models, allowing us to make stronger conclusions about the future utility of LLM simulation.

\section{Observational Scaling Laws for Predicting Downstream Utility}
\label{sec:obs_scaling_laws}

\textbf{Observational Scaling Laws.} In this section, we investigate how the current trajectory of LLM development will translate into utility for social scientists. The current trajectory of LLM development is focused on increasing LLMs' general intelligence. For this reason, we are interested in the relationship between LLM’s general benchmark performances and social science task performance. Despite the wide array of general intelligence tasks in existence, prior works have found a low-dimensional capability space can explain most of the model variation on general tasks \citep{ilic2024evidence,burnell2023revealing}. This common capability space allows direct comparison across families in a method known as observational scaling laws  \citep{ruan2024observational}. Unlike our experiments in \S\ref{sec:compute_scaling_laws}, which rely on a carefully-controlled suite of model checkpoints trained with fixed parameters on the same pre-training distribution, observational scaling laws allow us to evaluate larger and more capable models, which are often better-suited for social tasks \citep{ziems2024css}, but lack the controlled checkpoints to run standard compute scaling laws.

In addition to correlating general and social performance, observational evaluations allow us to predict downstream utility at scale. When evaluations reveal correlations between upstream metrics like loss and downstream metrics like accuracy, we can link our findings from \S\ref{sec:compute_scaling_laws} to predict future performance of models at much larger scales. These extrapolations assume the log-linear fit holds  beyond the fitted range and that the observational calibration function remains valid at orders of magnitude beyond observed models. While such extrapolations should be interpreted with caution, prior work demonstrates the viability of this approach. For example, \citet{held2026_delphi} was able to predict the loss of a 25B parameter model with just 0.2\% error by extrapolating their scaling laws $300\times$ past their largest run.

\paragraph{Models.} We evaluate a broad set of $35$ base LLMs from $7$ model families, including Gemma, Llama, OLMo, OPT, Phi, Qwen, and Yi. These models range from 0.5B to 70B parameters, and follow a variety of training recipes, from standard pre-training on web corpora to training with synthetic data. We opt for base models because we are interested in modeling heterogeneous distributions \citep{wu2025llm} for tasks like \texttt{WVS}, and post-trained models tend to collapse variance in models' learned opinion and behavior distributions \citep{hu2025simbench,sorensen2025spectrum}. 

\paragraph{General Benchmarks.} To represent the current paradigm in language modeling, we aggregate the performances of the above models on a wide range of general skills that model developers currently optimized for. These span instruction-following, world knowledge, commonsense, reasoning, and programming. Concretely, we consider \textbf{general knowledge} with (1) MMLU \citep{hendrycks2020measuring} and (2) MMLU-pro \citep{wang2024mmlu}; \textbf{instruction-following} with (3) IFEval \citep{zhou2023instruction}; \textbf{calibration} with (4) TruthfulQA \citep{lin2022truthfulqa}; \textbf{commonsense} with (5) HellaSwag \citep{zellers2019hellaswag}, (6) Winograd \citep{sakaguchi2021winogrande}, and XWinograd \citep{muennighoff2023crosslingual}; \textbf{mathematical reasoning} with (7) GSM8K \citep{cobbe2021training} and (8) MATH \citep{hendrycks2021measuring}; \textbf{knowledge-informed reasoning} with (9) GPQA \citep{rein2024gpqa} and (10) ARC-C \citep{clark2018think}; \textbf{multi-step reasoning} with (11) BIG-Bench Hard \citep[BBH;][]{suzgun2022challenging} and (12) MuSR \citep{sprague2023musr}; and \textbf{programming} with HumanEval \citep{chen2021evaluating}.

\paragraph{PC-1.} For each model, we project its general task performance into a single scalar, given by the first principal component of the PCA over its benchmarking results. Although we use different benchmarks than  \citet{ruan2024observational}, we similarly observe that PC-1 explains most of the variance (here 62\%).

\subsection{Results on \texttt{WVS}: Opinion Simulation}
\label{subsec:wvs_results}
\begin{figure}
    \centering
    \includegraphics[width=\linewidth]{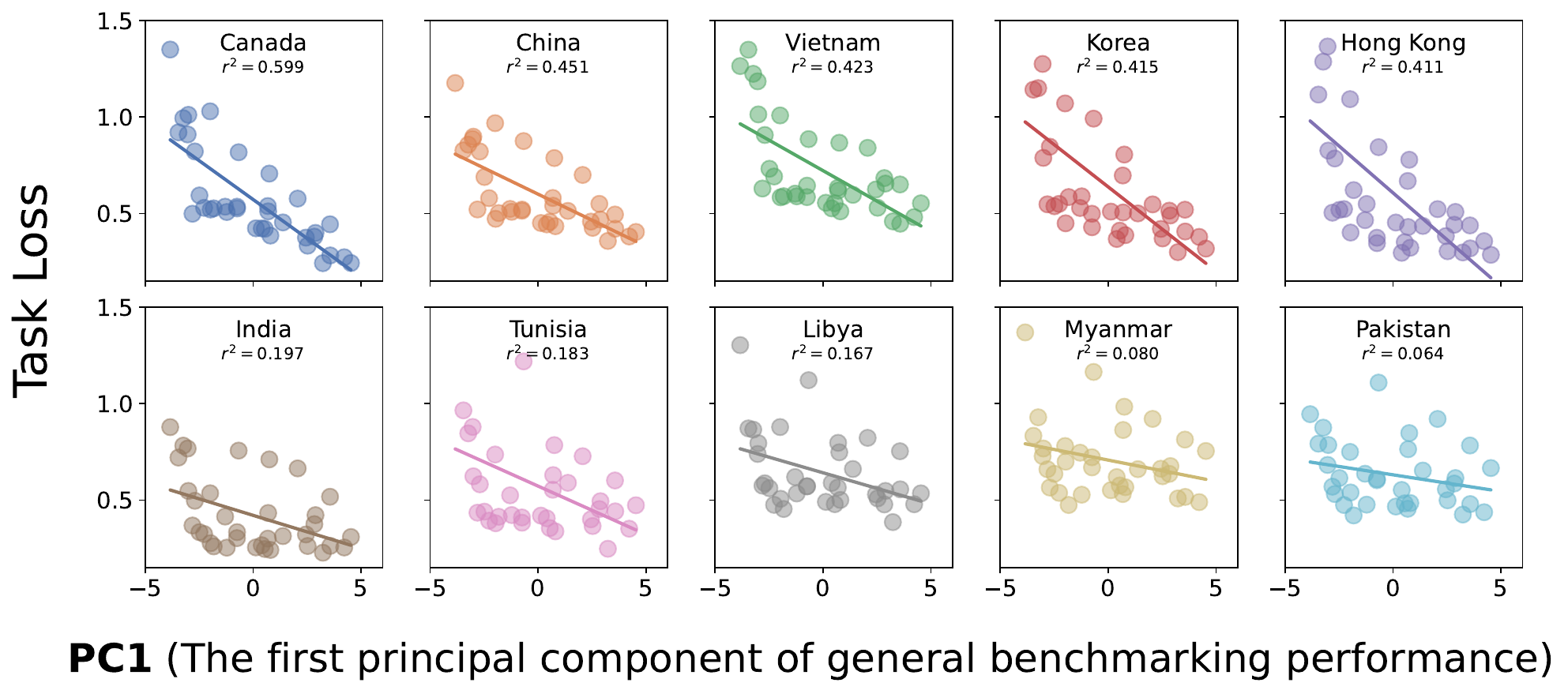}
    \caption{\small \textbf{Observational Scaling Laws for \texttt{WVS}}. \textbf{Top:} The five survey unit regions with the strongest correlation to general performance, $r^2>0.4$. \textbf{Bottom:} The five regions with the weakest correlation, $r^2<0.2$.} 
    \label{fig:obs_scaling_wvs}
\end{figure}

\textbf{Correlating General and Social Performance.} Opinion simulation is not uniformly aligned with general capabilities. Figure~\ref{fig:obs_scaling_wvs} shows the five regions most correlated (top) and five least correlated with general capabilities (bottom). Here we see that benchmarks explain as much as 60\% of simulation variance and as little as 6\%. The region with the least explained variance is Pakistan ($r^2=0.064$), followed by Myanmar ($r^2=0.080$): two populations that are under-represented in LLM pre-training data. Canada, a well-represented region, is most strongly correlated with general performance ($r^2=0.60$). 

\begin{wrapfigure}[23]{r}{0.38\textwidth}
    \vspace{-1.5em}
    \centering
    \includegraphics[width=\linewidth,clip]{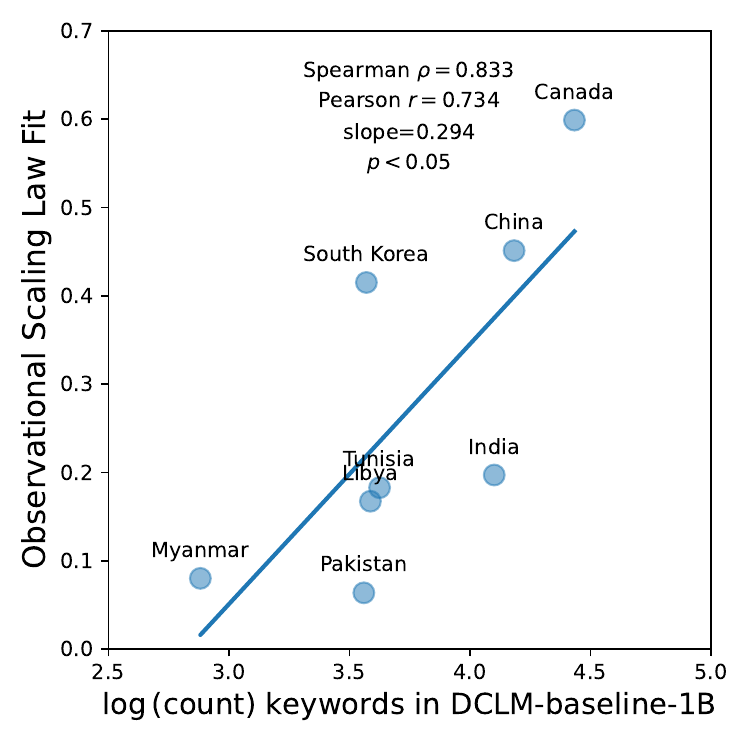}
    \caption{\small \textbf{Pre-training biases predict observational scaling laws.} We find a Spearman correlation of $\rho=0.8$ and a Pearson correlation of $r=0.7$ between observational fit and pre-training term frequency $(p<0.05)$, which supports our conclusions distributional imbalances in LLM pre-training data explain observational scaling discrepancies.} 
    \label{fig:dclm_analysis}
\end{wrapfigure}

The scaling discrepancies above are consistent with distributional imbalances in LLM pre-training data. We approximate distributional imbalances by counting region-specific keywords in a random 1B token sample of the DCLM pre-training corpus. Specifically, we count named references to the top 10 most populous metropolitan areas for each region and correlate the cumulative log frequency with the observational scaling law coefficients of determination. We find a Spearman correlation of $\rho=0.8$ and a Pearson correlation of $r=0.7$ between observational fit and pre-training term frequency $(p<0.05)$, which supports our conclusions about pre-training biases. For more details, see Appendix~\ref{appdx:pre-training_biases}.

Canadian simulation fidelity is intuitively most strongly correlated with commonsense reasoning tasks like Winograd ($r=-0.78$) and HellaSwag ($r=-0.73$), which measure implicit social reasoning and everyday expectations about behavior. Simulation fidelity is least correlated with  multi-step reasoning (MuSR $r=-0.42$), epistemic calibration (TruthfulQA $r=-0.47$), and programming abilities (HumanEval $r=-0.38$). From these results, we can conclude that opinion simulation is less about formal reasoning and understandably more about how people typically think in social situations. 

\textbf{Predicting Downstream Utility At Scale.}  For opinion simulation with the \texttt{WVS}, the task loss is the KL divergence between the survey results and the model's predicted opinion distribution. This is a directly interpretable metric without the need for correlations with other metrics like accuracy. Comparing across regions, observational models obtain a minimum average KL divergence of as high as 0.48 on the Myanmar subset and as low as 0.23 on the Indian subset. While these are still relatively high, introducing significant distributional distortion, our compute scaling results from \S\ref{sec:compute_scaling_laws} suggest that scale alone could substantially reduce KL divergence, approaching zero within a reasonable horizon. For the Myanmar scaling law (Figure~\ref{fig:compute_scaling_wvs}), the slope is -0.25 ($r^2=0.9$), so one can estimate the KL divergence will approach zero with as little as 83 times the compute.\footnote{Since $10^{\frac{0.48}{0.25}} = 83$} 

\subsection{Results on \texttt{Psych-101}: Behavioral Simulation}
\label{subsec:psych_101_obs}
\begin{figure}
    \centering
    \includegraphics[width=\linewidth]{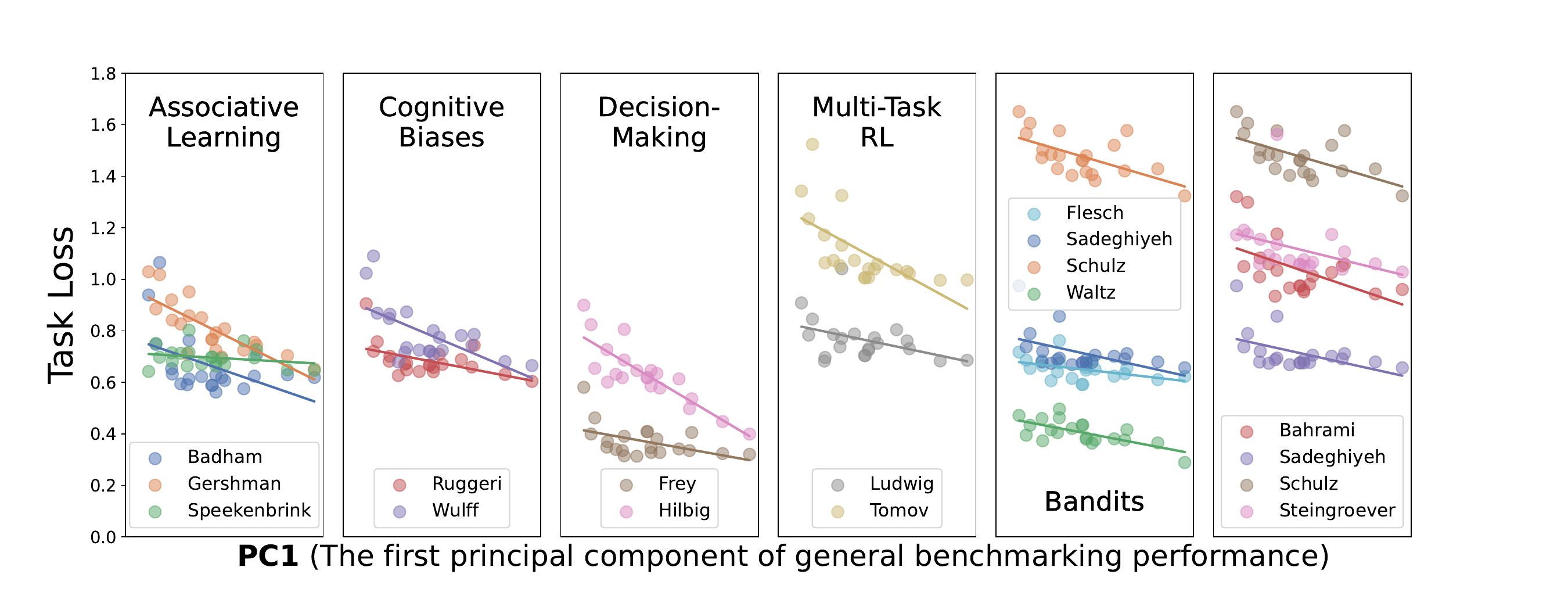}
    \caption{\small \textbf{Observational Scaling Laws for \texttt{Psych-101}} on representative subtasks. We cluster tasks by their experimental hypothesis domain, including \textit{associative learning} and \textit{cognitive biases}, but we do not observe any strong relationship between domain and scaling behavior.} 
    \label{fig:obs_scaling_psych_101}
\end{figure}

\textbf{Correlating General and Social Performance.} The alignment between behavioral simulation and general performance depends on the psychological experiment as shown in Figure~\ref{fig:obs_scaling_psych_101}. Most simulation subtasks are at least directionally correlated with general performance. However, one subtask from \citet{speekenbrink2008learning} appears to be entirely orthogonal to general performance, having a correlation and slope of zero (see the green line in the left subplot under \textit{Associative Learning}).

To understand which experiments are better explained by general performance, we clustered the \texttt{Psych-101} subtasks by domain: associative learning, cognitive biases, decision-making, multi-task RL, and reward maximizing bandits (see Figure~\ref{fig:obs_scaling_psych_101}). We found that neither the fit nor the slope appears to be a function of the psychological domain. Within associative learning, there is the subtask with zero slope, as well as a subtask from \citet{gershman2020origin} with the steepest slope and best fit ($r^2=0.63$). For a complete table of regression coefficients, see Table~\ref{tab:psych101_scaling_sorted} in Appendix~\ref{appdx:observational_scaling_laws}. Instead, slope and explained variance are functions of particular psychological experiments. 

In cases where general performance predicts behavioral simulation, task performance generally correlates with knowledge and reasoning. For example, in \citet{gershman2020origin}, performance most strongly correlates with the models' general knowledge (MMLU $r=-0.80$; MMLU Pro $r=-0.72$) and knowledge-informed reasoning (ARC-C $r=-0.74$), and it is least strongly correlated with instruction following (IFEval $r=-0.38$). 

\textbf{Predicting Downstream Utility At Scale.} Across subtasks, loss predicts accuracy sigmoidally or linearly (see Figure~\ref{fig:loss_accuracy_psych_101} in Appendix~\ref{appdx:prompts}). A small set of subtasks are nearing performance saturation with accuracies above 90\%. On other unsolved subtasks, LLMs appear capable of improving within a reasonable scale horizon. For example, \citet{gershman2020origin} has a maximum accuracy of 72\%, from Qwen2.5 14B with a loss of 0.7. To reach 90\% accuracy would require dropping the loss to 0.2, following the linear fit in Figure~\ref{fig:loss_accuracy_psych_101}. Using the scaling fit from \S\ref{sec:compute_scaling_laws} (Figure~\ref{fig:compute_scaling_psych101}), we can determine that such a performance gain would require approximately 40 times more compute.

On the other hand, scale alone is \textit{not} expected to improve at least a third of the behavioral simulation subtasks, one of which has no observed compute-scaling behavior \citep{schulz2020finding}, and seven of which have weak correlation ($r^2<0.3$) between accuracy and task loss \citep{flesch2018comparing,frey2017cct,hebart2023things, ludwig2023human,plonsky2018when,sadeghiyeh2020temporal,wulff2018description}. Generally, these tasks have the weakest observational slopes and do not align well with general model capabilities, thus indicating that they at least partially occupy a capability space orthogonal to math and reasoning. We investigate these tasks further in \S\ref{sec:weak_scaling}.

\subsection{Results on \texttt{ACL}: Longitudinal Forecasting}
\begin{figure}
    \centering
    \includegraphics[width=0.9\linewidth]{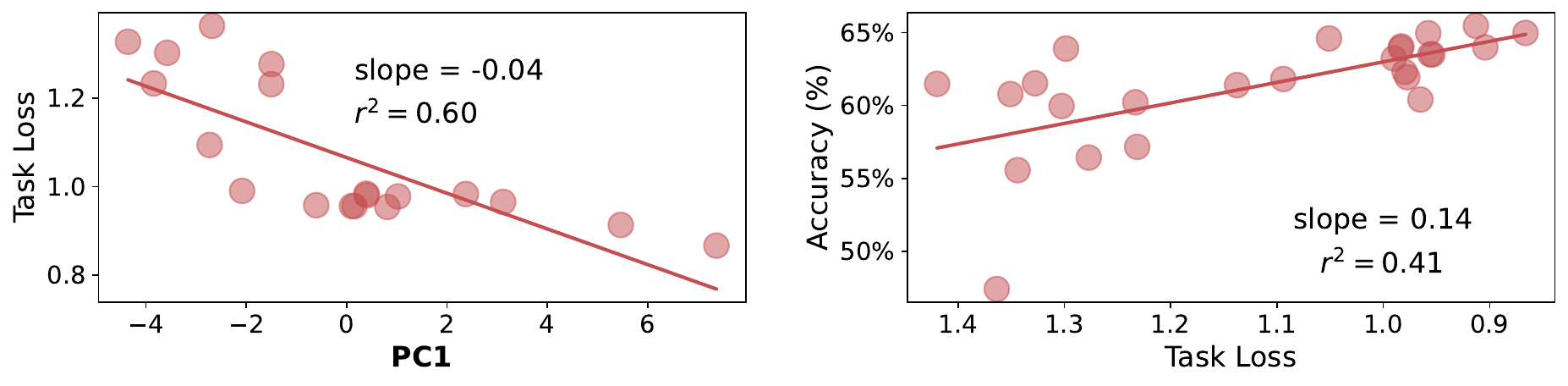}
    \caption{\small \textbf{Observational Scaling and Downstream Forecasting for \texttt{ACL}.} \textbf{Left:} Observational scaling law with task loss. \textbf{Right:} We show that loss linearly correlates with accuracy.} 
    \label{fig:obs_scaling_longitudinal}
\end{figure}

\textbf{Correlating General and Social Performance.} Results on the left side of Figure~\ref{fig:obs_scaling_longitudinal} show that longitudinal forecasting improves with generally more capable models (slope = -0.04), and that performance on this task is correlated with general benchmarking performance ($r^2 = 0.60$). Longitudinal forecasting is most strongly correlated with epistemic calibration ($r=-0.43$), general knowledge (MMLU Pro $r=-0.36$), and instruction-following (IFEval $r=-0.34$), and least strongly correlated with multi-step reasoning (MuSR $r=-0.04$) and programming abilities (HumanEval $r=-0.10$).

\textbf{Predicting Downstream Utility At Scale.} The plot on the right side reveals a linear relationship between task loss and accuracy. Combined with the clean compute-scaling results on this task (\S\ref{sec:compute_scaling_laws}), this suggests that the current trajectory in large language modeling will produce tools that can more capably forecast the results of longitudinal studies. However, the rate of improvement may be prohibitively slow. The best model we evaluated observationally was Yi 1.5 34B, which achieved an accuracy of 65\%, with 0.9 loss. Our composite scaling estimates suggest that improving performance by an additional 5\% on this task would require loss to fall by 0.36, demanding models trained optimally on over 1000 times more compute. As Yi 1.5 34B was already trained on 3.6T tokens, we can quickly see the bottleneck. Moreover, if the current trends continue with scale, we would expect an upper-bound model accuracy of only 77\%. Together, these results suggest that longitudinal forecasting will not be solved by scaling pre-training compute alone.

\subsection{Takeaways from Observational Scaling}

Together, the results above show that general model capabilities roughly predict simulation fidelity for three representative inference problems: the prediction of social states, behavioral trajectories, and longitudinal variables. Here, we find that models with the strongest knowledge and knowledge-informed reasoning capabilities are also the most calibrated simulators. On the other hand, programming and formal multi-step reasoning abilities are less correlated with simulation fidelity. Moreover, general capabilities do not uniformly translate into better social simulators. Simulation fidelity is a function of pre-training data representation. The evidence here suggests that under-represented populations will be underserved by the current scaling paradigm. 

\section{Making Sense of Scaling Laws for Behavior Simulation}
\label{sec:weak_scaling}

\subsection{Characterizing Weaker Scaling Effects}
\label{subsec:characterize_weak_scaling}
The results in prior sections are optimistic and surprising. We observe that scaling up compute (\S\ref{sec:compute_scaling_laws}) and general model capabilities  (\S\ref{sec:obs_scaling_laws}) alone will produce LLMs that are generally more capable of faithfully simulating human opinions, behavior, and longitudinal wellbeing. This effect is continuous and has an observable distribution, where most tasks demonstrably scale to some degree. In demonstrating this effect, we have also narrowed the space of potential simulation tasks which may not scale. Most notably, about a third of the behavior simulation tasks have weak calibration fits ($r^2<0.3$) between loss and downstream accuracy. 
Table~\ref{tab:psych_101_strong_weak_tasks} in the Appendix sorts behavior simulation tasks by the strength of the compute scaling law fit and accuracy calibration fit. As discussed in \S\ref{subsec:psych_101_obs}, the effect of scale is not immediately predictable from the high-level task domain, since, for example, \textit{multi-armed bandits} have both strong (\citeauthor{gershman2020origin}) and weak effects (\citeauthor{sadeghiyeh2020temporal}, \citeauthor{schulz2020finding}). However, within-domain analysis allows us to isolate other explanatory factors.

\textbf{Multi-armed bandits.} All four of the best-scaling tasks in Table~\ref{tab:psych_101_strong_weak_tasks} \citep{wilson2014humans,gershman2020origin,wu2018generalisation,gershman2018deconstructing} are bandit problems with independent and stationary reward distributions. Like humans, large models can readily infer these rewards via Rescorla-Wagner-style conditioning \citep{rescorla1972theory}, and thus faithfully simulate behavior. In contrast, \citet{schulz2020finding}, at the very bottom of the table, is a bandit problem with correlated reward distributions. Participants and models may have different heuristics for inferring the latent correlations and generalizing across tasks, leading to the poor scaling calibration. Similarly, models are uncalibrated in the non-stationary bandit of \citet{bahrami2020four}. Second, a related computational challenge is catastrophic forgetting, where LLMs and human solutions differ in \citet{flesch2018comparing}. Third, we observe poor scaling for bandit tasks across individual differences like trait temporal discounting in \citet{sadeghiyeh2020temporal}, or mental illness in \citet{waltz2020differential}, which may be due to heterogeneity in the population that skews away from what the model is exposed to in pre-training data. 

\textbf{Associative Learning.} In this domain, \citet{badham2017deficits} is a category learning tasks that scales nicely, whereas \citep{speekenbrink2008learning} is poorly calibrated. The primary difference between these tasks is that the latter studies individual learning differences between individuals with amnesia and those without. As in the multi-armed bandits, so also here we see that population skew may be a challenge for LLM scaling.

\textbf{Multi-task RL.} The strongest-scaling multi-task RL problem is from \citet{tomov2021multitask}, while the weakest-scaling problem is from \citet{ludwig2023human}. Both study how humans can generalize by learning from related tasks. The difference is that \citet{ludwig2023human} modifies the congruence between related tasks as an experimental variable, so that similar tasks may require very different policies. This too introduces heterogeneity which is not present in the easier \citet{tomov2021multitask}.  

\textbf{Cognitive Biases.} Studies of risk-aversion are notoriously idiosyncratic, since individuals' prospect theory parameters can vary across sessions \citep{glockner2012cognitive}. This may partly explain the weaker loss calibration in risk aversion studies \citet{plonsky2018when} and \citet{peterson2021using}, relative to other cognitive biases like \citet{wulff2018description}.

\subsection{Parameter Scaling with Fine-tuned Models}
\label{sec:finetuning}

In \S\ref{subsec:characterize_weak_scaling} we observed the following common features across behavior simulation tasks with weaker scaling: (1) \textbf{structured reward spaces} that require generalization (e.g., \citeauthor{schulz2020finding}), (2) \textbf{catastrophic forgetting} (e.g., \citeauthor{flesch2018comparing}), (3) \textbf{population heterogeneity} and skew (e.g., \citeauthor{sadeghiyeh2020temporal}), (4) \textbf{task heterogeneity} (e.g., \citeauthor{ludwig2023human}), and (5) \textbf{individual idiosyncrasies} (e.g., \citeauthor{plonsky2018when}). The five tasks listed here include the four with the weakest calibration fit \citep{flesch2018comparing,ludwig2023human,plonsky2018when,sadeghiyeh2020temporal}, as well as \citet{schulz2020finding}, the only task with a scaling coefficient of zero. 

Fine-tuning can help us better diagnose the limitations of scale in the tasks above. The problem may be that in-context learning emerges for these tasks only beyond particular scale thresholds \citep{wei2022emergent,liu2022few}. We expect that fine-tuning can produce learning signals earlier than in-context-learning, particularly in the domain of social simulation \citep{kolluri2025finetuning}. Moreover, scaling laws can be highly sensitive to design decisions like prompt formatting \citep{sclar2024quantifying, held2025relative}, and fine-tuning can reduce the sensitivity of our analysis to prompt design. Finally, fine-tuning can possibly address the model's misalignment to population and task heterogeneity by fitting to the target population and task distributions. 

\textbf{Fine-tuning.} We fine-tune Qwen2.5 (\citeyear{qwen25_technical_report}) and Llama3 (\citeyear{llama3}), which have base model checkpoints at multiple parameter scales. Specifically, we evaluate Qwen2.5 \{0.5B, 1.5B, 3B, 7B\} and Llama 3 \{1B, 3B, 8B\}. Beyond this scale, we confront optimization issues due to the scarcity of our training data. We fine-tune each model on 16M tokens from the full Psych-101 dataset with a learning rate sweep to select the best configuration. See Appendix~\ref{appdx:finetuning} for more training details. Finally, we the parameter scaling behavior of our fine-tuned models on these weakest scaling tasks. As a control, we also evaluate on two strong compute scaling tasks: \citet{gershman2020origin}, which had strongest scaling fit ($r^2=0.97)$, and \citet{hilbig2014generalized}, which had the best calibration ($r^2=0.86$).

\begin{figure}
    \centering
    \includegraphics[width=\linewidth]{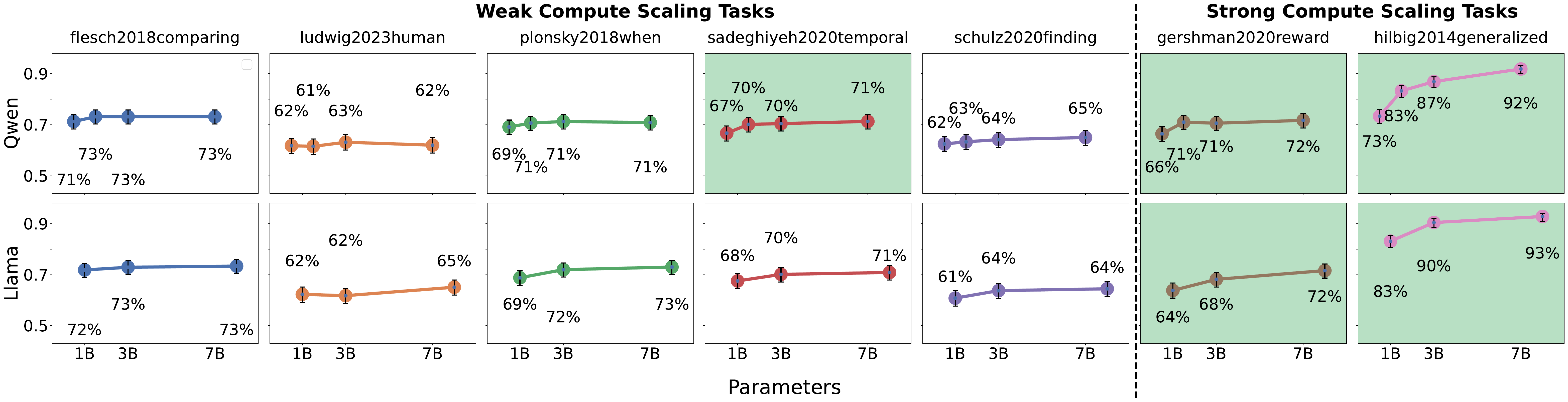}
    \caption{\small \textbf{Parameter Scaling after Finetuning on Psych-101.} We finetune Qwen2.5 and Llama3 at different scales on each task respectively. Highlighted \hlgreen{green} are experiments in which the largest model's advantage over the smallest model is statistically significant ($p<0.05$). \textbf{Left:} For tasks with weak compute scaling, we observe little or no evidence of parameter scaling. \textbf{Right:} For tasks with strong compute scaling, we also observe evidence of parameter scaling.
    } 
    \label{fig:finetuning}
\end{figure}

\paragraph{Results.} Figure~\ref{fig:finetuning} shows parameter scaling curves after finetuning. Highlighted in \hlgreen{green} are experiments in which the largest model's advantage over the smallest model is statistically significant ($p<0.05$). The five columns on the left show results from the five tasks with the weakest compute scaling. Importantly, we observe statistically significant parameter scaling in \citet{sadeghiyeh2020temporal}, which supports our hypothesis that, under population skew and  heterogeneity, distributional alignment can help strengthen weak in-context scaling curves.\footnote{We might expect the same result then for task heterogeneity in \citet{ludwig2023human}. However, the fine-tuned model performance on \citet{ludwig2023human} is not even monotonically increasing with parameter scale, so we will refrain from interpreting the null result as anything more than task noise.} In other words, weak scaling here is likely a product of distributional misalignment, and not evidence that the behavior is orthogonal or somehow outside the domain of what models can represent.

Second, we fail to observe statistically significant parameter scaling in either Qwen2.5 or Llama3 on any of \citeauthor{flesch2018comparing}, \citeauthor{ludwig2023human}, \citeauthor{plonsky2018when}, or \citeauthor{schulz2020finding}. One might argue these flat curves are a shortcoming of fine-tuning in a low-data regime. However, the right side of Figure~\ref{fig:finetuning} discredits the argument, since two related demonstrate statistically significant parameter scaling in the same low-data training regime. This means the flat fine-tuning curves on the left are diagnostic of the tasks rather than the fine-tuning recipe. These null results allow us to more confidently assert that the weak scaling phenomenon is unlikely to be a thresholding effect of in-context learning. Instead, we argue that, over the interval tested in this experiment, scale alone appears insufficient to close the social simulation gap where human heuristics differ from models in domains like structured reward spaces, catastrophic forgetting, and individual idiosyncrasies.

\section{Conclusion}
This work asks whether the current trajectory of LLM development will produce increasingly more faithful models for social and experimental simulation. To answer this, we ran carefully-controlled compute scaling law experiments, and established accuracy-calibration functions from observational experiments to link compute scaling with downstream performance. Together, these experiments suggested that, in the aggregate, compute scaling is a consistent and reliable lever for progress in social simulation over finite probability spaces, especially when modeling populations that well-represented in pre-training corpora. This result is surprising, given the many known limitations of LLM simulations \citep{bisbee2024synthetic,bowers2025centaur,tan2024language}. With this in mind, we believe it is wise for scientists of well-represented populations to begin investing in LLM-driven survey simulations for applications like pilot studies \citep{anthis2025llm}, because with rapid progress, it will be cheaper, faster, and more reliable to simulate than to survey participants in regions like Canada and China. 

Conversely, for underrepresented populations, LLM survey simulations will not as rapidly improve, and if LLMs are widely and injudiciously adopted for social simulation, this disparity in relative scaling laws \citep{held2025relative} may ultimately exacerbate the \textit{generalizability crisis} in the social sciences \citep{yarkoni2022generalizability}. Observational experiments show that opinion simulations for Pakistan and Myanmar, and some behavioral simulation tasks like \citet{flesch2018comparing} occupy a capability space that is at least partially distinct from LLMs' general knowledge and reasoning abilities. These distinct tasks have the weakest observational slopes, and are generally those for which models fail to translate scale to downstream performance. For such tasks, even fine-tuned Qwen2.5 and Llama3 models fail to demonstrate performance improvements with scale, indicating that compute scaling alone is insufficient to improve performance for this small set of social simulation tasks.

\section*{Limitations and Future Work}

\textbf{Task Scope.} Our conclusions about social simulation cover only tasks with small, discrete action spaces because likelihood-based loss is well-defined here. The tasks we selected represent three of the most common MCQ-style inference problems in social simulation. Our conclusions should be seen as a careful exploration of scaling in its most natural simulation domains rather than an exhaustive treatment of scaling across the simulation literature. Future work can apply observational scaling analysis to the simulation of open-ended surveys, multi-turn dialogue, and multi-agent interaction where the target is a semantic property of an entire sequence or interaction protocol rather than a next-token distribution.

\textbf{Model Scope and Capabilities.} We evaluate only base models, which allow us to both isolate the effects of pre-training compute scale on simulation fidelity and to avoid the issue of mode collapse in post-trained models. Our conclusions do not preclude the benefits of post-training for social simulation. Pre-training compute scale alone is not expected to predict the simulation fidelity of frontier reasoning models, or models with retrieval augmented generation and tool use. We provide clean insights into compute scaling, but to better understand the future of social simulation, future work should consider the effect of such additional modeling levers as well. 

\textbf{Model Size and Extrapolation.} Our compute scaling laws are fit over two orders of magnitude, from $10^{18}$ to $10^{20}$ FLOPs, which is standard in controlled scaling analyses \citep{mcleish2026gemstones,held2025relative,biderman2023pythia}. The fitted range is limited by the computational cost of pre-training a suite of 80+ IsoFLOP checkpoints. Our downstream utility predictions in \S\ref{sec:obs_scaling_laws} extrapolate on these fits, and we assume the log-linear fit will hold past our largest run. Although prior work has shown these extrapolations can be accurate at $300\times$ with as little as 0.2\% error \citet{held2026_delphi}, one should still interpret our extrapolations with caution as magnitude effect-size estimates rather than precise forecasts.

\textbf{Simulation Fidelity and Ethics.} Faithful LLM simulations should be calibrated to the ground truth distributions of the human populations they simulate. Our findings show that simulation fidelity is, in part, a function of biases in pre-training data, so we caution against the use of LLMs in simulating underrepresented populations. Moreover, fidelity is a necessary but not sufficient for scientifically valid and ethically responsible simulation. Our work measures only distributional fidelity, not scientific or ethical validity. Social scientists have expressed numerous concerns around transparency \citep{fecher2025friend}, bias \citep{schroeder2025qualitative}, and ethics \citep{liao2024research} of LLM simulations, and we encourage practitioners to carefully consider these concerns before deployment.


\section*{Acknowledgments}
This work was supported by the Google TPU Research Cloud (TRC), a Stanford HAI–GCP Grant, a Stanford HAI Seed Grant, as well as by DSO National Laboratories. The authors thank Myra Cheng, Joe Baumann, Ananya Bhattacharjee, and the SALT Lab broadly for discussions and feedback during the development of this work.

\bibliography{main}
\bibliographystyle{colm2026_conference}

\newpage
\appendix
\onecolumn

\section{Prompts, Task Design, and Evaluation Protocols}
\label{appdx:prompts}

\begin{figure}
    \centering
    \includegraphics[width=\linewidth]{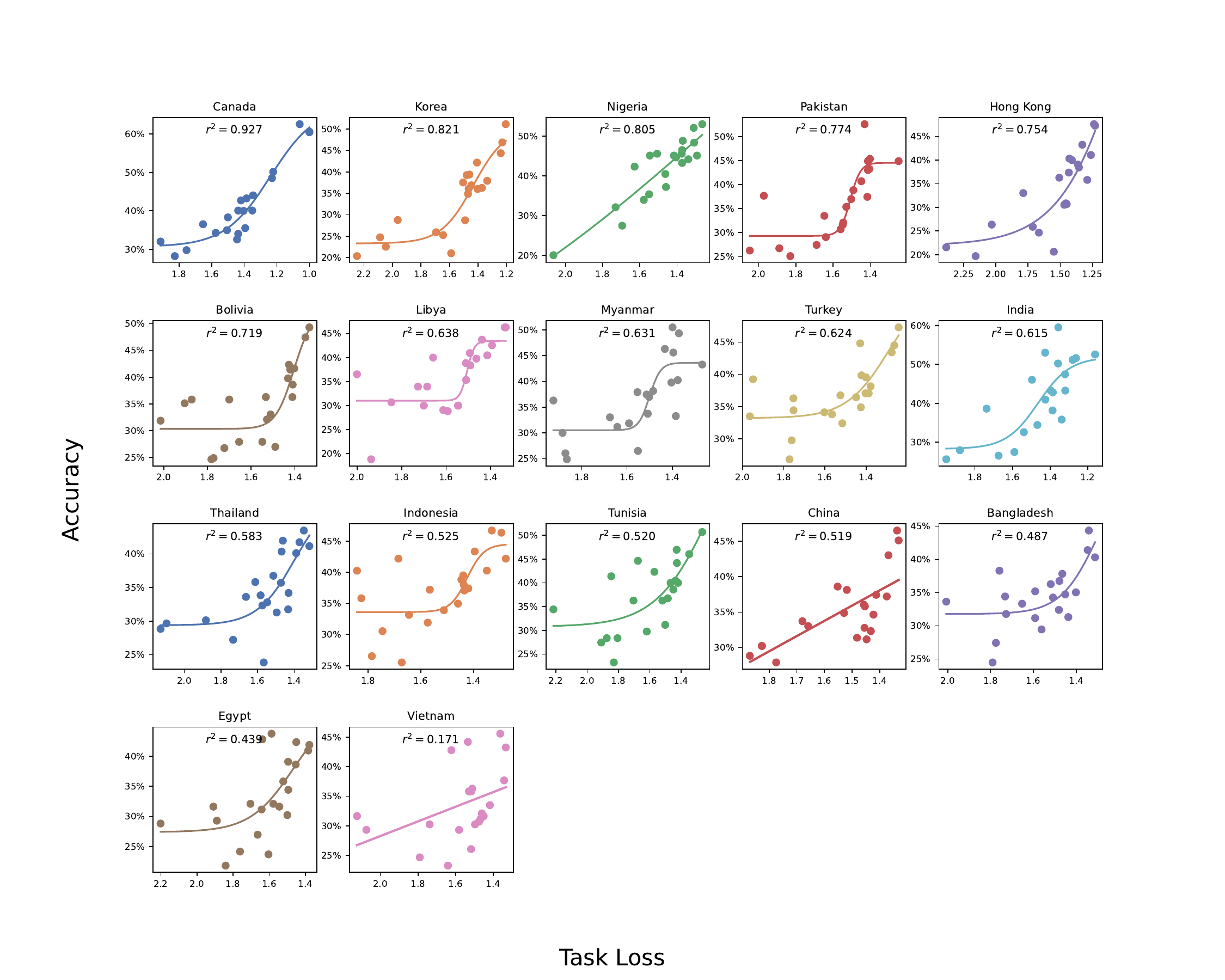}
    \caption{\textbf{The Model's Negative Log Likelihood (NLL) on the Majority Answer Generally Predicts Accuracy Sigmoidally} for \texttt{WVS} tasks ($r^2>0.4$ for 16 out of 17 tasks)} 
    \label{fig:loss_accuracy_wvs}
\end{figure}

\begin{figure}
    \centering
    \includegraphics[width=\linewidth]{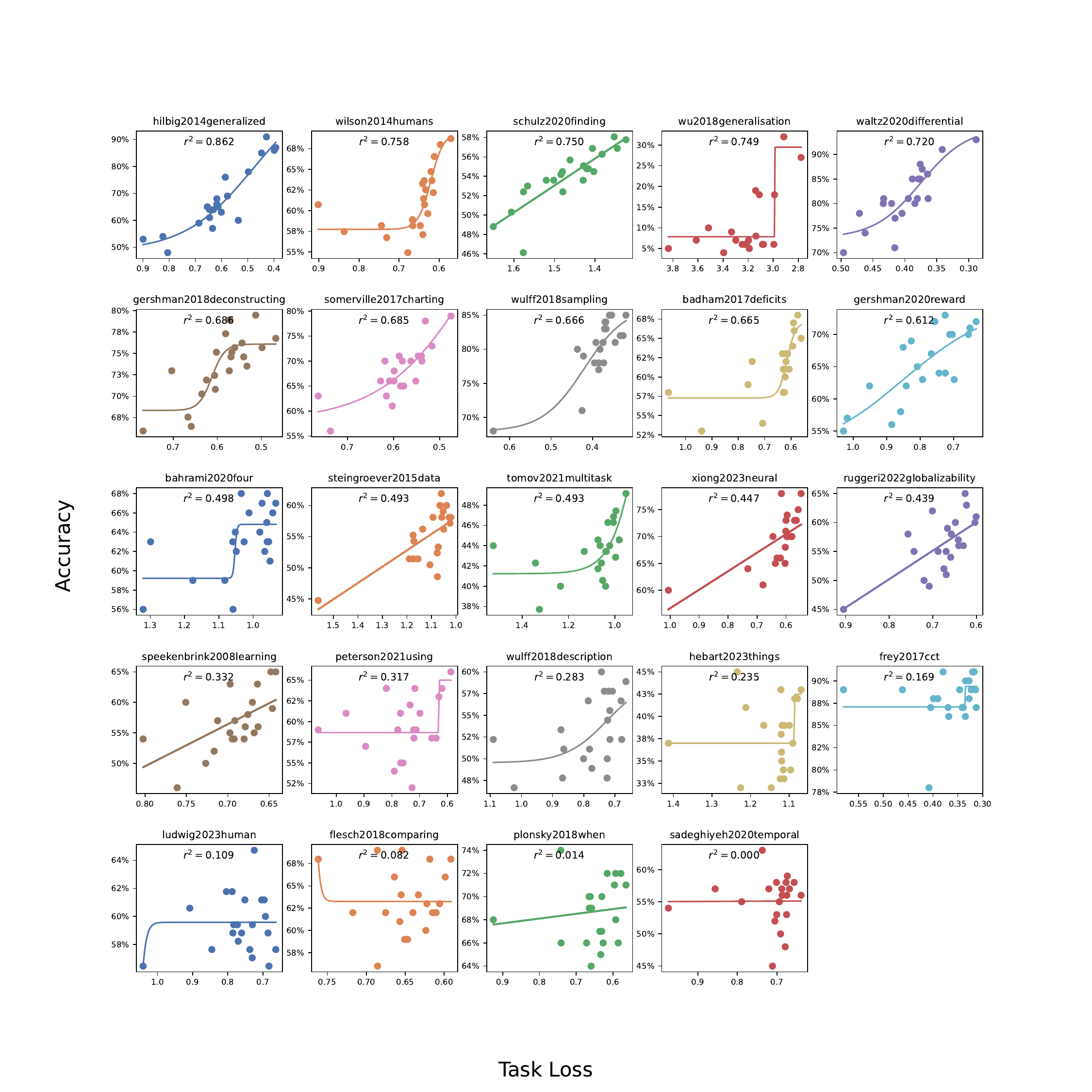}
    \caption{\textbf{The Model's Negative Log Likelihood (NLL) on the Correct Answer Generally Predicts Accuracy Sigmoidally} for the majority of \texttt{Psych-101} tasks ($r^2>0.4$ for 15 out of 24 tasks)} 
    \label{fig:loss_accuracy_psych_101}
\end{figure}

Scaling laws can be highly sensitive to the manner in which tasks are operationalized, prompts are formatted, and metrics are defined. With respect to metrics, raw log-probabilities generally scale smoothly, while metrics like accuracy can appear more erratic due to thresholding effects \citep{mirage, difficult}. However, scaling experiments with loss are meaningful only if they can predict downstream metrics of interest. For this reason, we have worked to optimize our tasks and prompts such that loss can generally predict accuracy. As shown in Figure~\ref{fig:loss_accuracy_wvs}, loss predicts \texttt{WVS} accuracy sigmoidally with $r^2>0.4$ for 16 out of 17 regions. In Figure~\ref{fig:loss_accuracy_psych_101} we see that loss predicts \texttt{Psych-101} accuracy sigmoidally with $r^2>0.4$ for 15 out of 24 psychological experiments.

\subsection{\texttt{WVS}}
\label{appdx:prompt_wvs}

Here, we evaluate the similarity between gold and predicted distributions using KL divergence. We choose not to use accuracy measures here for a few reasons. First, opinion simulations can involve imbalanced outcomes or small treatment effects. Models can often achieve high accuracy on these tasks by predicting the base rates or prior distributions for social variables (i.e., predicting only the most dominant opinion). Models that optimize for accuracy metrics may also underperform for minority groups. In short, distributional similarity helps account for distributional variance, skew, and tail behavior.

To reach align loss and accuracy for this task, we followed best practices from \citet{held2025relative} and formatted \texttt{WVS} as a multiple-choice (MCQ) task with letter labels and options, and the probabilities are computed over the full label+option strings.

An example of a prompt for this task is as follows:

\begin{lstlisting}
Region: Bangladesh

Question: What is your biological sex?
A. Male
B. Female
C. N/A

Response:
B. Female

Question: Based on the year of your birth, in which generational group do you fall?
A. Silent Generation (1928 - 1945)
B. Baby Boomer (1946 - 1964)
C. Generation X (1965 - 1980)
D. Millennial (1981 - 1996)
E. Generation Z (1997 - 2012)
F. Generation Alpha (2013 - 2025)
G. N/A

Response:
D. Millennial (1981 - 1996)

Question: What is the highest educational level that you have attained?
A. Primary education
B. Lower secondary education
C. Upper secondary education
D. Post-secondary non-tertiary education
E. Short-cycle tertiary education
F. Bachelor or equivalent
G. Master or equivalent
H. Doctoral or equivalent
I. N/A

Response:
A. Primary education

Question: Which of the following socioeconomic classes do you believe you belong to?
A. Upper class
B. Upper middle class
C. Lower middle class
D. Working class
E. Lower class
F. N/A

Response:
D. Working class

Question: How important is religion in your life?
A. Very important
B. Rather important
C. Not very important
D. Not at all important
E. Don't know

Response:
A. Very important

Question: How would you describe the type of settlement where this interview is being conducted?
A. Urban
B. Rural
C. N/A

Response:
B. Rural

Question: How would you rate the importance of family in your life?
A. Very important
B. Rather important
C. Not very important
D. Not at all important

Response:
\end{lstlisting}

\subsection{\texttt{Psych-101}}
\label{appdx:prompt_psych101}
An example of a prompt for this task is as follows:

\begin{lstlisting}
You will be asked to repeatedly choose between four different options labeled T, Y, P, and Z. You select an option by pressing the corresponding key on your keyboard. Each time you select an option, you will get a different number of points. Your goal is to win as many points as possible.

From the options ("T", "Y", "P", "Z"), you press "T" and get 80.0 points.
From the options ("T", "Y", "P", "Z"), you press "Y" and get 81.0 points.
From the options ("T", "Y", "P", "Z"), you press "P" and get 61.0 points.
From the options ("T", "Y", "P", "Z"), you press "
\end{lstlisting}

\subsection{\texttt{ACL}}
\label{appdx:prompt_acl}
Our methodology for identifying salient outcome, explanatory, and control variables relies on the hundreds of publications that make use of the ACL. Specifically, we scrape PDF manuscripts for 379 studies that rely on ACL data and use GPT-4.1-mini to parse these studies and determine which ACL variables can be considered outcomes, explanatory, or control variables. We keep as targets any variables used as outcomes in at least one study. We define that outcome's controls and explanatory variables as the minimum set of controls and explanatory variables from \textit{any} paper that considered the target outcome.

An example of a prompt for this task is as follows:

\begin{lstlisting}
(1) Did your (husband/wife) get a high school diploma or pass a high school equivalency test? (Options: "yes", "no", "inappropriate", "don't know", "na")

In the year 1989: "yes"

(2) How often do you usually attend religious services? (Options: "more than once a week", "once a week", "2 or 3 times a month", "about once a month", "less than once a month", "never", "don't know", "na")

In the year 1989: "less than once a month"
In the year 1994: "less than once a month"
In the year 2001: "about once a month"
In the year 2011: "less than once a month"
In the year 2019: "less than once a month"

(3) In a typical week, about how many times do you talk on the telephone with friends, neighbors or relatives? (Options: "more than once a day", "once a day", "2 or 3 times a week", "about once a week", "less than once a week", "never or no phone", "don't know", "na")

In the year 1989: "2 or 3 times a week"
In the year 1994: "2 or 3 times a week"
In the year 2001: "once a day"
In the year 2011: "once a day"
In the year 2019: "
\end{lstlisting}

\section{Additional Compute Scaling Laws}
\label{appdx:compute_scaling_laws}
\begin{figure}
    \centering
    \includegraphics[width=\linewidth]{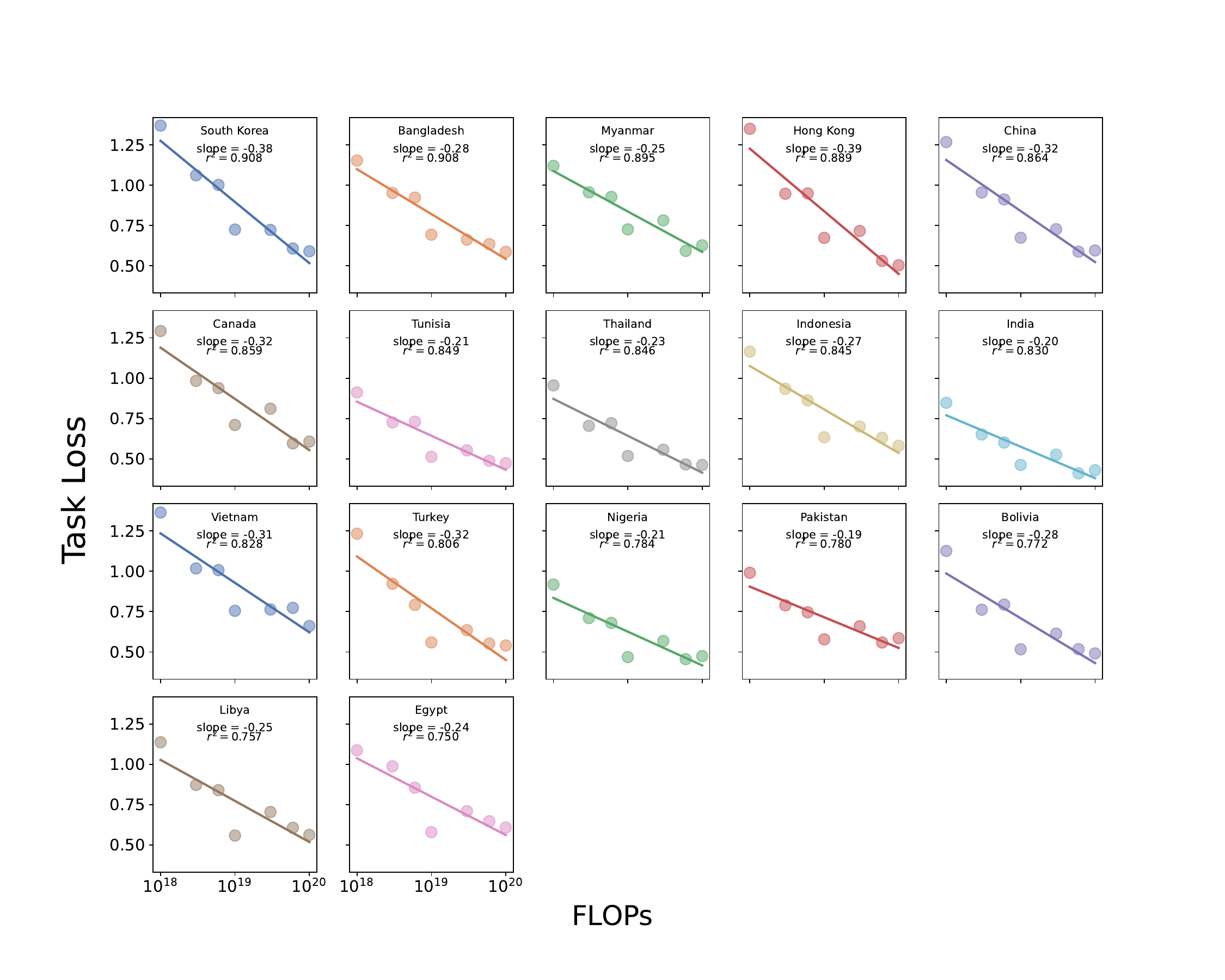}
    \caption{\textbf{Compute Scaling Laws for \texttt{WVS}.} We observe log-linear improvements in task loss on all \texttt{WVS} subtasks by scaling compute alone with models trained on DCLM \citep{dclm} from $10^{18}$ to $10^{20}$ FLOPs.} 
    \label{fig:compute_scaling_wvs}
\end{figure}

\begin{figure}
    \centering
    \includegraphics[width=\linewidth]{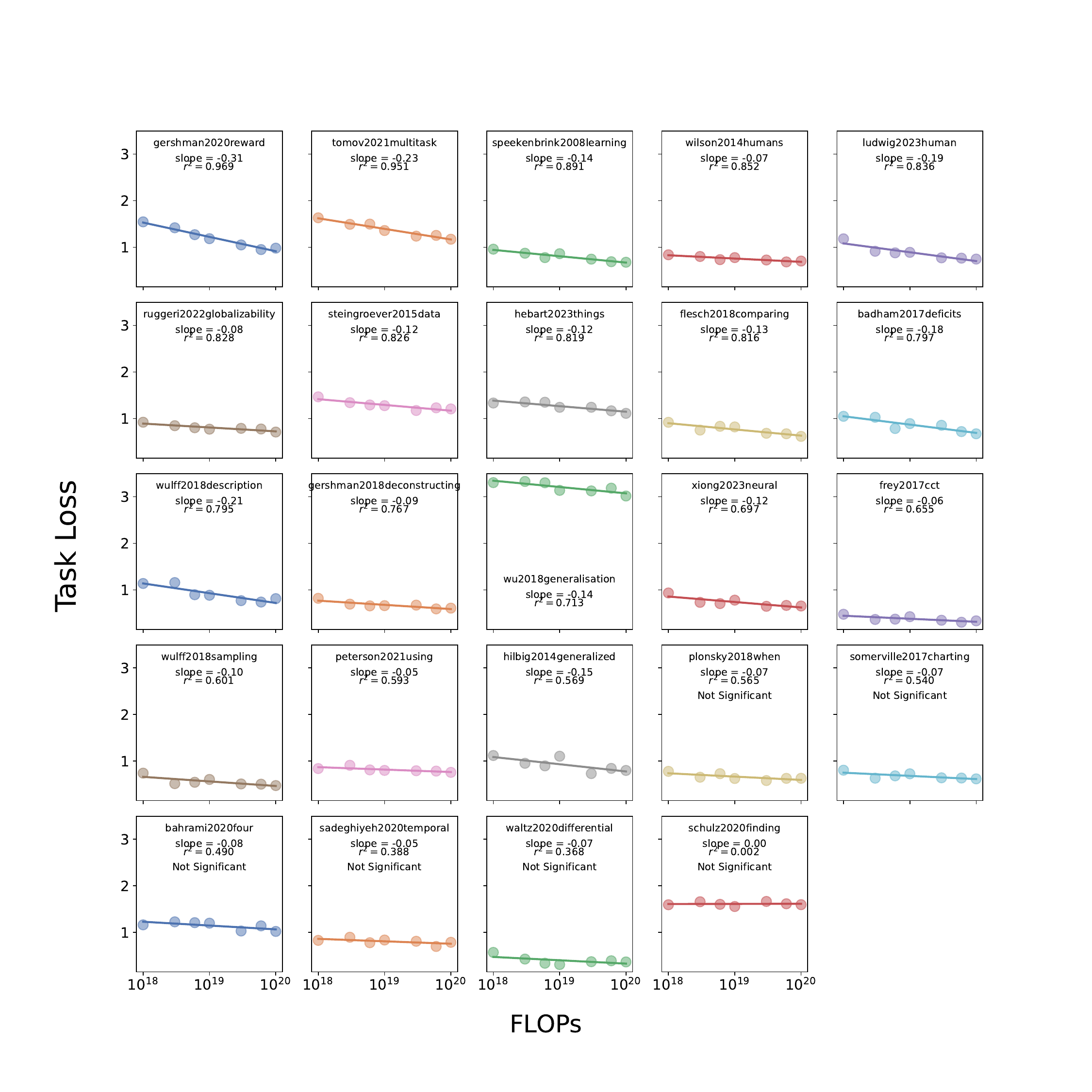}
    \caption{\textbf{Compute Scaling Laws for \texttt{Psych-101}.} We observe log-linear improvements in task loss on all \texttt{Psych-101} subtasks by scaling compute alone with models trained on DCLM \citep{dclm} from $10^{18}$ to $10^{20}$ FLOPs. However, the slope is not significant with $p<0.05$ for \textbf{\citet{plonsky2018when}, somerville2017charting, bahrami2020four, \citet{sadeghiyeh2020temporal}, \citet{waltz2020differential}}, and \textbf{\citet{schulz2020finding}}.} 
    \label{fig:compute_scaling_psych101}
\end{figure}

We provide detailed plots of compute scaling laws for each subtask of \texttt{WVS} in Figure~\ref{fig:compute_scaling_wvs}, and each subtask of \texttt{Psych-101} in Figure~\ref{fig:compute_scaling_psych101}.

\section{Additional Observational Scaling Laws}
\label{appdx:observational_scaling_laws}
Table~\ref{tab:psych101_scaling_sorted} provides additional details on our observational scaling law fits for the \texttt{Psych-101} subtasks, sorted by $r^2$ (descending).

The list of model cards we evaluated observationally includes: 01-ai/Yi-1.5-34B, 01-ai/Yi-1.5-9B, 01-ai/Yi-34B, 01-ai/Yi-6B, Qwen/Qwen1.5-0.5B, Qwen/Qwen1.5-1.8B, Qwen/Qwen1.5-14B, Qwen/Qwen1.5-4B, Qwen/Qwen1.5-7B, Qwen/Qwen2-0.5B, Qwen/Qwen2-1.5B, Qwen/Qwen2-7B, allenai/OLMo-1B-hf, allenai/OLMo-7B-hf, bigscience/bloom-7b1, facebook/opt-1.3b, facebook/opt-13b, facebook/opt-2.7b, facebook/opt-30b, facebook/opt-6.7b, facebook/xglm-1.7B, facebook/xglm-4.5B, facebook/xglm-7.5B, google/gemma-2-27b, google/gemma-2-2b, google/gemma-2b, google/gemma-7b, huggyllama/llama-13b, huggyllama/llama-7b, meta-llama/Llama-2-13b-hf, meta-llama/Llama-2-70b-hf, meta-llama/Llama-2-7b-hf, meta-llama/Meta-Llama-3.1-8B, microsoft/phi-1\_5, microsoft/phi-2

\begin{table}[t]
\centering
\small
\begin{tabular}{llllll}
\toprule
\textbf{Subtask} & \textbf{Category} & \textbf{Slope} & \textbf{$r^2$} & \textbf{$r$} & \textbf{p-value}  \\
\midrule
\citet{hilbig2014generalized} & Decision-Making & -0.03 & 0.72 & 0.85 & 0.00 \\
\citet{gershman2020origin} & Reward Maximization & -0.03 & 0.63 & 0.79 & 0.00 \\
\citet{waltz2020differential} & Single-Arm Bandit & -0.01 & 0.47 & 0.69 & 0.00 \\
\citet{peterson2021using} & Cognitive Biases & -0.02 & 0.46 & 0.68 &  0.00 \\
\citet{tomov2021multitask} & Multi-Task RL & -0.03 & 0.43 & 0.66 & 0.00 \\
\citet{wulff2018description} & Cognitive Biases & -0.02 & 0.41 & 0.64 & 0.00 \\
\citet{schulz2020finding} & Multi-Arm Bandit & -0.02 & 0.36 & 0.60 & 0.00 \\
\citet{somerville2017charting} & Reward Maximization & -0.01 & 0.33 & 0.57 & 0.01 \\
\citet{wilson2014humans} & Reward Maximization & -0.01 & 0.31 & 0.56 & 0.01 \\
\citet{bahrami2020four} & Multi-Arm Bandit & -0.02 & 0.28 & 0.53 &  0.01 \\
\citet{ruggeri2022globalizability} & Cognitive Biases & -0.01 & 0.26 & 0.51 & 0.02 \\
\citet{sadeghiyeh2020temporal} & Single-Arm Bandit & -0.01 & 0.24 & 0.49 &  0.02 \\
\citet{frey2017cct} & Decision-Making & -0.01 & 0.23 & 0.48 & 0.03 \\
\citet{badham2017deficits} & Associative Learning & -0.02 & 0.21 & 0.46 & 0.04 \\
\citet{flesch2018comparing} & Single-Arm Bandit & -0.01 & 0.21 & 0.46 & 0.04 \\
\citet{hebart2023things} & Misc & -0.01 & 0.18 & 0.42 & 0.06 \\
\citet{ludwig2023human} & Multi-Task RL & -0.01 & 0.17 & 0.41 & 0.06 \\
\citet{steingroever2015data} & Reward Maximization & -0.01 & 0.13 & 0.36 & 0.11 \\
\citet{plonsky2018when} & Cognitive Biases & -0.01 & 0.12 & 0.35 & 0.13 \\
\citet{wulff2018sampling} & Reward Maximization & -0.01 & 0.12 & 0.35 & 0.12 \\
\citet{speekenbrink2008learning} & Associative Learning & -0.00 & 0.06 & 0.24 & 0.31 \\
\citet{wu2018generalisation} & Multi-Arm Bandit & 0.01 & 0.02 & 0.14 & 0.58 \\
\bottomrule
\end{tabular}
\caption{Observational scaling law fits across Psych-101 subtasks, sorted by $r^2$ (descending).}
\label{tab:psych101_scaling_sorted}
\end{table}

\section{Pre-training Biases Predict Observational Scaling Laws}
\label{appdx:pre-training_biases}

Here, we investigate whether the scaling discrepancies in Figure~\ref{fig:obs_scaling_wvs} are consistent with distributional imbalances in LLM pre-training data. To do so, we correlate the strength of observational fits with the log frequency of region-related terms in the DCLM-baseline, which is the pre-training corpus that was used to train our compute scaling law suite (\S\ref{sec:compute_scaling_laws}). We analyze a statistically unbiased random sample of 1B tokens from the DCLM-baseline, prepared by \citet{sharma2025billion}. As a proxy for pre-training prevalence, we count the frequency of terms corresponding to the most populous metropolitan areas in each region. To avoid confounds, we exclude any ambiguous metropolitan area names. For example, while \textit{Hyderabad} is the 7th largest city in Pakistan, it could be confused with \textit{Hyderabad, India}; therefore we exclude it. After exclusion, we have a list of 10 city names for each region. City terms and counts for each region are given in Table~\ref{tab:region_terms}.

\begin{table}[t]
\centering
\small
\begin{tabular}{rlp{7cm}p{3cm}}
\toprule
\textbf{Region} & \textbf{Count} & \textbf{Cities} & \textbf{Excluded}  \\
\midrule
Canada & 27,107 & \textit{Toronto, Montreal, Calgary, Ottawa, Winnipeg, Mississauga, Vancouver, Quebec City, Markham, Gatineau}  & \textit{Edmonton, Brampton, Hamilton, Surrey} \\
China & 15,235 & \textit{Shanghai, Beijing, Chongqing, Tianjin, Guangzhou, Shenzhen, Chengdu, Wuhan, Dongguan, Hangzhou} & N/A\\
India & 12,601 & \textit{Mumbai, Delhi, Bengaluru, Ahmedabad, Pune, Surat, Jaipur, Lucknow, Kanpur, Nagpur} & \textit{Hyderabad, Chennai}\\
Tunisia & 4,215 & \textit{Tunis, Sfax, Sousse, Kairouan, Bizerte, Gabes, Ariana, Gafsa, Monastir, Ben Arous} & N/A\\
Libya & 3,862 & \textit{Tripoli, Benghazi, Misrata, Zawiya, Bayda, Ajdabiya, Tobruk, Sabha, Khoms, Derna}& N/A\\
South Korea & 3,717 & \textit{Seoul, Busan, Incheon, Daegu, Daejeon, Gwangju, Suwon, Ulsan, Changwon, Yongin} & N/A \\
Pakistan & 3,627 & \textit{Karachi, Lahore, Faisalabad, Rawalpindi, Gujranwala, Peshawar, Multan, Islamabad, Quetta, Bahawalpur} & \textit{Hyderabad} \\
Myanmar & 761 & \textit{Yangon, Mandalay, Naypyidaw, Mawlamyine, Bago, Pathein, Monywa, Meiktila, Myingyan, Taunggyi} & N/A\\
\bottomrule
\end{tabular}
\caption{\textbf{City terms and counts for each region} in a 1B token sample of the DCLM-baseline. Excluded city terms are those which are ambiguous, such as \textit{Hyderabad} which could refer to the city in India or the city in Pakistan.}
\label{tab:region_terms}
\end{table}

In Figure~\ref{fig:dclm_analysis}, we correlate the log frequency of city terms in the DCLM with the observational scaling law fit. We find a Spearman correlation of $\rho=0.8$ and a Pearson correlation of $r=0.7$ between observational fit and pre-training term frequency $(p<0.05)$, which supports our conclusions distributional imbalances in LLM pre-training data explain observational scaling discrepancies.

\section{Fine-tuning Details}
  \label{appdx:finetuning}

  We fine-tune Qwen2.5 models at five parameter scales (0.5B, 1.5B, 3B, 7B) on each of our social simulation tasks. All models are initialized from HuggingFace checkpoints and trained with a next-token prediction objective over the task-formatted text.

  \paragraph{Training Configuration.} We train each model for 1000 steps with a global batch size of 4 and a sequence length of 4096, yielding approximately 16M tokens per run.  For each
  model--task pair, we sweep learning rates over $\{1\times 10^{-5}, 3\times 10^{-5},3 \times 10^{-4}\}$ and select the configuration with the lowest validation loss. The 0.5B--7B models are trained on TPU v5p-8.

  \begin{table}[h]
  \centering
  \small
  \begin{tabular}{lrr}
  \toprule
  \textbf{Model} & \textbf{Tokens} & \textbf{FLOPs} \\
  \midrule
  Qwen2.5-0.5B & 144M & $4.3 \times 10^{17}$ \\
  Qwen2.5-1.5B & 144M & $1.3 \times 10^{18}$ \\
  Qwen2.5-3B  & 144M & $2.6 \times 10^{18}$ \\
  Qwen2.5-7B  & 144M & $6.0 \times 10^{18}$ \\
  \midrule
  Total        & 576M & $1.0 \times 10^{19}$ \\
  \bottomrule
  \end{tabular}
  \caption{Fine-tuning compute budget per model across all tasks and learning rate configurations (3 tasks $\times$ 3 learning rates $\times$ 16M tokens per run). Estimated via $6ND$.}
  \label{tab:finetuning_compute}
  \end{table}

  \textbf{Splits.} For \texttt{Psych-101}, we create train and test splits of 49K and 520 experimental transcripts (144M and 2M tokens respectively), assuring that there is no overlap in the participants from these sets as a contamination check. 

\begin{table}[tbp]
\centering
\resizebox{\columnwidth}{!}{%
\begin{tabular}{l l p{6.5cm} c c c c}
\toprule
\textbf{Reference} & \textbf{Domain} & \textbf{Original Study Question} & \textbf{Participant Recruitment} & \textbf{Scaling Fit} & \textbf{Calibration Fit} & \textbf{Mean Fit} \\
\midrule
\citet{wilson2014humans} & Reward Maximization & How does the time horizon of a task impact how humans navigate the explore-exploit dilemma? & Princeton students & 0.85 & 0.76 & \hl{\textbf{0.80}}\\
\addlinespace
\citet{gershman2020origin} & Reward Maximization & How do people navigate the explore-exploit dilemma? & MTurk & \hl{\textbf{0.97}} & 0.61 & 0.77 \\
\addlinespace
\citet{wu2018generalisation} & Multi-armed bandit & How do humans navigate the explore-exploit dilemma? & MTurk & 0.71 & 0.75 & 0.73 \\
\addlinespace
\citet{gershman2018deconstructing} & Multi-armed bandit & How do humans navigate the explore-exploit dilemma? & MTurk & 0.77 & 0.69 & 0.73\\
\addlinespace
\citet{badham2017deficits} & Associative Learning & How do people solve category learning problems? & Unspecified & 0.80 & 0.66 & 0.73 \\
\addlinespace
\citet{hilbig2014generalized} & Decision-making & How do consumers make purchasing decisions using reviews with different levels of trustworthiness? & Unspecified & 0.57 & \hl{\textbf{0.86}} & 0.70 \\
\addlinespace
\citet{tomov2021multitask} & Multi-task RL & How do people transfer knowledge to solve related tasks? & MTurk & 0.95 & 0.49 & 0.68 \\
\addlinespace
\citet{steingroever2015data} & Reward Maximization & How do humans navigate the explore-exploit dilemma? & Unspecified & 0.83 & 0.49 & 0.64 \\
\addlinespace
\citet{somerville2017charting} & Reward Maximization & How does age impact people's strategies for navigating the explore-exploit dilemma? & Unspecified & 0.54 & 0.69 & 0.61 \\
\addlinespace
\citet{ruggeri2022globalizability} & Cognitive Biases & What are the differences in how people from 61 different countries delay gratification? & Email, Social Media & 0.83 & 0.44 & 0.60 \\
\addlinespace
\citet{speekenbrink2008learning} & Associative Learning & How do individuals with amnesia learn to solve a category learning problems? & Korsakoff's syndrome & 0.89 & 0.33 & 0.54 \\
\addlinespace
\citet{waltz2020differential} & Multi-armed bandit & How does schizophrenia impact people's strategies for navigating the explore-exploit dilemma? & UMD Med School & 0.37 & 0.72 & 0.52\\
\addlinespace
\citet{bahrami2020four} & Multi-armed bandit & How do people navigate the explore-exploit dilemma with a \textit{restless bandit} (where the rewards drift over time)? & Unspecified & 0.49 & 0.50 & 0.49\\
\addlinespace
\citet{wulff2018description} & Cognitive Biases & How does the description-experience gap predict human decision making? & Meta-Analysis & 0.80 & 0.29 & 0.48 \\
\addlinespace
\citet{hebart2023things} & Analytical Thinking & How do fMRIs explain the human ability to identify the odd-one-out in a set? & MTurk & 0.82 & 0.24 & 0.44\\
\addlinespace
\citet{peterson2021using} & Cognitive Biases & How do cognitive biases like risk aversion influence people's risk assessment (following prospect theory)? & MTurk & 0.59 & 0.32 & 0.43 \\
\addlinespace
\citet{frey2017cct} & Decision-Making  & Is there a general factor of risk preference? & Meta-Analysis &  0.66 & 0.17 & 0.33 \\
\addlinespace
\citet{ludwig2023human} & Multi-task RL & How does the congruence between related tasks influence people's ability to learn those tasks? & Prolific & 0.84 & 0.11 & 0.30 \\
\addlinespace
\citet{flesch2018comparing} & Multi-armed bandit & How do humans perform continual learning without catastrophic forgetting? & MTurk & 0.82 & 0.08 & 0.26 \\
\addlinespace
\citet{plonsky2018when} & Cognitive biases & How do cognitive biases like risk aversion influence people's risk assessment (following prospect theory)? & Unspecified & 0.57 & 0.01 & 0.08 \\
\addlinespace
\citet{sadeghiyeh2020temporal} & Multi-armed bandit & Does temporal discounting explain how people navigate the explore-exploit dilemma? & Arizona Psych students & 0.39 & 0.00 & 0.00 \\
\addlinespace
\citet{schulz2020finding} & Multi-armed bandit & How do humans use generalization to maximize rewards over correlated multi-armed bandits? & MTurk & 0.00 & 0.75 & 0.00 \\
\bottomrule
\end{tabular}
}
\caption{\small \textbf{Psych-101 Tasks sorted by the geometric mean between the compute scaling law fit and the accuracy calibration fit.} Best fits are \hl{\textbf{highlighted}}. We observe some common features across with weaker scaling tasks: (1) structured reward spaces that require generalization (e.g., \citeauthor{schulz2020finding}), (2) catastrophic forgetting (e.g., \citeauthor{flesch2018comparing}), (3) population heterogeneity and skew (e.g., \citeauthor{sadeghiyeh2020temporal}), (4) task heterogeneity (e.g., \citeauthor{ludwig2023human}), and (5) individual ideosyncracies (e.g., \citeauthor{plonsky2018when}). 
}
\label{tab:psych_101_strong_weak_tasks}
\end{table}

\end{document}